%% file: main_header.tex
\documentclass[runningheads]{llncs}

\usepackage[T1]{fontenc}

\usepackage{graphicx}
\usepackage{amsmath}
\usepackage{enumitem}
\usepackage{float}
\usepackage[font=footnotesize]{subfig}
\usepackage{color}
\usepackage{verbatim}
\usepackage{tikz}
\usepackage{makecell}
\usepackage{xurl}
\usepackage{booktabs}
\usepackage{algorithm}
\usepackage{algorithmic}

\usepackage[misc]{ifsym}
\usepackage[hidelinks]{hyperref}

\usepackage{fancyhdr}
\fancypagestyle{firststyle}
{
   \fancyhf{}
   \fancyfoot[C]{\footnotesize \thepage}
   \fancyhead[C]{\small To appear in Proceedings of ECML PKDD 2022}
}

\begin{document}

\title{No More Strided Convolutions or Pooling:\\A New CNN Building Block for Low-Resolution Images and Small Objects}
\toctitle{No More Strided Convolutions or Pooling: A New CNN Building Block for Low-Resolution Images and Small Objects}
\titlerunning{SPD-Conv for Low-Resolution Images and Small Objects}

\tocauthor{Raja~Sunkara and Tie~Luo}
\authorrunning{R. Sunkara and T. Luo}
\author{Raja Sunkara
\and
Tie Luo
\thanks{Corresponding author}\textsuperscript{\Letter}}

\institute{Computer Science Department, Missouri University of Science and Technology \\
\email{\{rs5cq,tluo\}@mst.edu}}

\maketitle 
\thispagestyle{firststyle}

\begin{abstract}
Convolutional neural networks (CNNs) have made resounding success in many computer vision tasks such as image classification and object detection. However, their performance degrades rapidly on tougher tasks where images are of low resolution or objects are small. In this paper, we point out that this roots in a defective yet common design in existing CNN architectures, namely the use of {\em strided convolution} and/or {\em pooling layers}, which results in a loss of fine-grained information and learning of less effective feature representations. To this end, we propose a new CNN building block called {\em SPD-Conv} in place of each strided convolution layer and each pooling layer (thus eliminates them altogether). SPD-Conv is comprised of a {\em space-to-depth} (SPD) layer followed by a {\em non-strided} convolution (Conv) layer, and can be applied in most if not all CNN architectures. We explain this new design under two most representative computer vision tasks: object detection and image classification. We then create new CNN architectures by applying SPD-Conv to YOLOv5 and ResNet, and empirically show that our approach significantly outperforms state-of-the-art deep learning models, especially on tougher tasks with low-resolution images and small objects. We have open-sourced our code at \url{ https://github.com/LabSAINT/SPD-Conv}.
\end{abstract}

\section{Introduction}

Since AlexNet \cite{krizhevsky2012imagenet}, convolutional neural networks (CNNs) have excelled at many computer vision tasks. 
For example in image classification, well-known CNN models include AlexNet, VGGNet ~\cite{simonyan2014very}, ResNet ~\cite{he2016deep}, etc.; while in object detection, those models include the R-CNN series~\cite{girshick2015fast,ren2016faster}, YOLO series~\cite{redmon2018yolov3,bochkovskiy2020yolov4}, SSD~\cite{liu2016ssd}, EfficientDet~\cite{tan2020efficientdet}, and so on. However, all such CNN models need ``good quality'' inputs (fine images, medium to large objects) in both training and inference. For example, AlexNet was originally trained and evaluated on $227 \times 227$ clear images, but after reducing the image resolution to 1/4 and 1/8, its classification accuracy drops by 14\% and 30\%, respectively \cite{koziarski2018impact}. The similar observation was made on VGGNet and ResNet too \cite{koziarski2018impact}. In the case of object detection, SSD suffers from a remarkable mAP loss of 34.1 on 1/4 resolution images or equivalently 1/4 smaller-size objects, as demonstrated in \cite{haris2021task}. In fact, small object detection is a very challenging task 
because smaller objects inherently have lower resolution, and also limited context information for a model to learn from. Moreover, they often (unfortunately) co-exist with large objects in the same image, which (the large ones) tend to dominate the feature learning process, thereby making the small objects undetected. 

In this paper, we contend that such performance degradation roots in a defective yet common design in existing CNNs. That is, the use of strided convolution and/or pooling, especially in the earlier layers of a CNN architecture. The adverse effect of this design usually does not exhibit because most scenarios being studied are ``amiable'' where images have good resolutions and objects are in fair sizes; therefore, there is plenty of {\em redundant} pixel information that strided convolution and pooling can {\em conveniently skip} and the model can still learn features quite well. However, in tougher tasks when images are blurry or objects are small, the lavish assumption of redundant information no longer holds and the current design starts to suffer from loss of fine-grained information and poorly learned features.

To address this problem, we propose a new building block for CNN, called {\em SPD-Conv}, in substitution of (and thus eliminate) strided convolution and pooling layers altogether. SPD-Conv is a {\em space-to-depth} (SPD) layer followed by a {\em non-strided} (i.e., vanilla) convolution layer. The SPD layer downsamples a feature map $X$ but retains all the information in the {\em channel} dimension, and thus there is no information loss. We were inspired by an image transformation technique \cite{sajjadi2018frame} which rescales a raw image before feeding it into a neural net, but we substantially generalize it to downsampling {\em feature maps} inside and throughout the entire network; furthermore, we add a non-strided convolution operation after each SPD to reduce the (increased) number of channels using learnable parameters in the added convolution layer. Our proposed approach is both {\em general} and {\em unified}, in that SPD-Conv  (i) can be applied to most if not all CNN architectures and (ii) replaces both strided convolution and pooling the same way. In summary, this paper makes the following contributions:
\begin{enumerate}[label={\arabic*)}]
\item We identify a defective yet common design in existing CNN architectures and propose a new building block called SPD-Conv in lieu of the old design. SPD-Conv downsamples feature maps without losing learnable information, completely jettisoning strided convolution and pooling operations which are widely used nowadays.

\item SPD-Conv represents a general and unified approach, which can be easily applied to most if not all deep learning based computer vision tasks.

\item Using two most representative computer vision tasks, object detection and image classification, we evaluate the performance of SPD-Conv. Specifically, we construct YOLOv5-SPD, ResNet18-SPD and ResNet50-SPD, and evaluate them on COCO-2017, Tiny ImageNet, and CIFAR-10 datasets in comparison with several state-of-the-art deep learning models. The results demonstrate significant performance improvement in AP and top-1 accuracy, especially on small objects and low-resolution images. See Fig.~\ref{fig:example} for a preview.

\item SPD-Conv can be easily integrated into popular deep learning libraries such as PyTorch and TensorFlow, potentially producing greater impact. Our source code is available at \url{ https://github.com/LabSAINT/SPD-Conv}.
\end{enumerate}

The rest of this paper is organized as follows. Section 2 presents background and reviews related work. Section 3 describes our proposed approach and Section 4 presents two case studies using object detection and image classification. Section 5 provides performance evaluation. This paper concludes in Section 6.

\input{tables/comparison_plot}

\section{Preliminaries and Related Work}\label{sec:relwk}

We first provide an overview for this area, focusing more on object detection since it subsumes image classification.

Current state-of-the-art object detection models are CNN-based and can be categorized into one-stage and two-stage detectors, or anchor-based or anchor-free detectors. 
A two-stage detector firstly generates coarse region proposals and secondly classifies and refines each proposal using a head (a fully-connected network). In contrast, a one-stage detector skips the region proposal step and runs detection directly over a dense sampling of locations. Anchor-based methods use {\em anchor boxes}, which are a predefined collection of boxes that match the widths and heights of objects in the training data, to improve loss convergence during training. We provide Table \ref{table:cls} that categorizes some well-known models.

Generally, one-stage detectors are faster than two-stage ones and anchor-based models are more accurate than anchor-free ones. Therefore, later in our case study and experiments we focus more on one-stage and anchor-based models, i.e., the first cell of Table \ref{table:cls}.

\input{tables/try}

A typical one-stage object detection model is depicted in Fig.~\ref{fig:object_de}. It consists of a CNN-based {\em backbone} for visual feature extraction and a detection {\em head} for predicting class and bounding box of each contained object. In between, a {\em neck} of extra layers is added to combine features at multiple scales to produce semantically strong features for detecting objects of different sizes.

\subsection{Small Object Detection}

Traditionally, detecting both small and large objects is viewed as a multi-scale object detection problem. A classic way is {\em image pyramid} \cite{adelson1984pyramid}, which resizes input images to multiple scales and trains a dedicated detector for each scale. To improve accuracy, SNIP~\cite{singh2018analysis} was proposed which performs {\em selective backpropagation} based on different object sizes in each detector. SNIPER~\cite{singh2018sniper} improves the efficiency of SNIP by only processing the context regions around each object instance rather than every pixel in an image pyramid, thus reducing the training time. Taking a different approach to efficiency, Feature Pyramid Network (FPN) \cite{lin2017feature} exploits the multi-scale features inherent in convolution layers using lateral connections and combine those features using a top-down structure. Following that, PANet \cite{liu2018path} and BiFPN \cite{tan2020efficientdet} were introduced to improve FPN in its feature information flow by using shorter pathways. Moreover, SAN \cite{kim2018san} was introduced to map multi-scale features onto a scale-invariant subspace to make a detector more robust to scale variation. All these models unanimously use strided convolution and max pooling, which we get rid of completely.

\subsection{Low-Resolution Image Classification}

One of the early attempts to address this challenge is \cite{chevalier2015lr}, which proposes an end-to-end CNN model by adding a super-resolution step before classification. Following that, \cite{peng2016fine} proposes to transfer fine-grained knowledge acquired from high-resolution training images to low-resolution test images. However, this approach requires high-resolution training images corresponding to the specific application (e.g., the classes), which are not always available.

This same requirement of high-resolution training images is also needed by several other studies such as \cite{wang2016studying}. Recently, \cite{singh2021enhancing} proposed a loss function that incorporate attribute-level separability (where attribute means fine-grained, hierarchical class labels) so that the model can learn class-specific discriminative features. However, the fine-grained (hierarchical) class labels are difficult to obtain and hence limit the adoption of the method.

\begin{figure}[t]
    \centering
    \includegraphics[width=\linewidth]{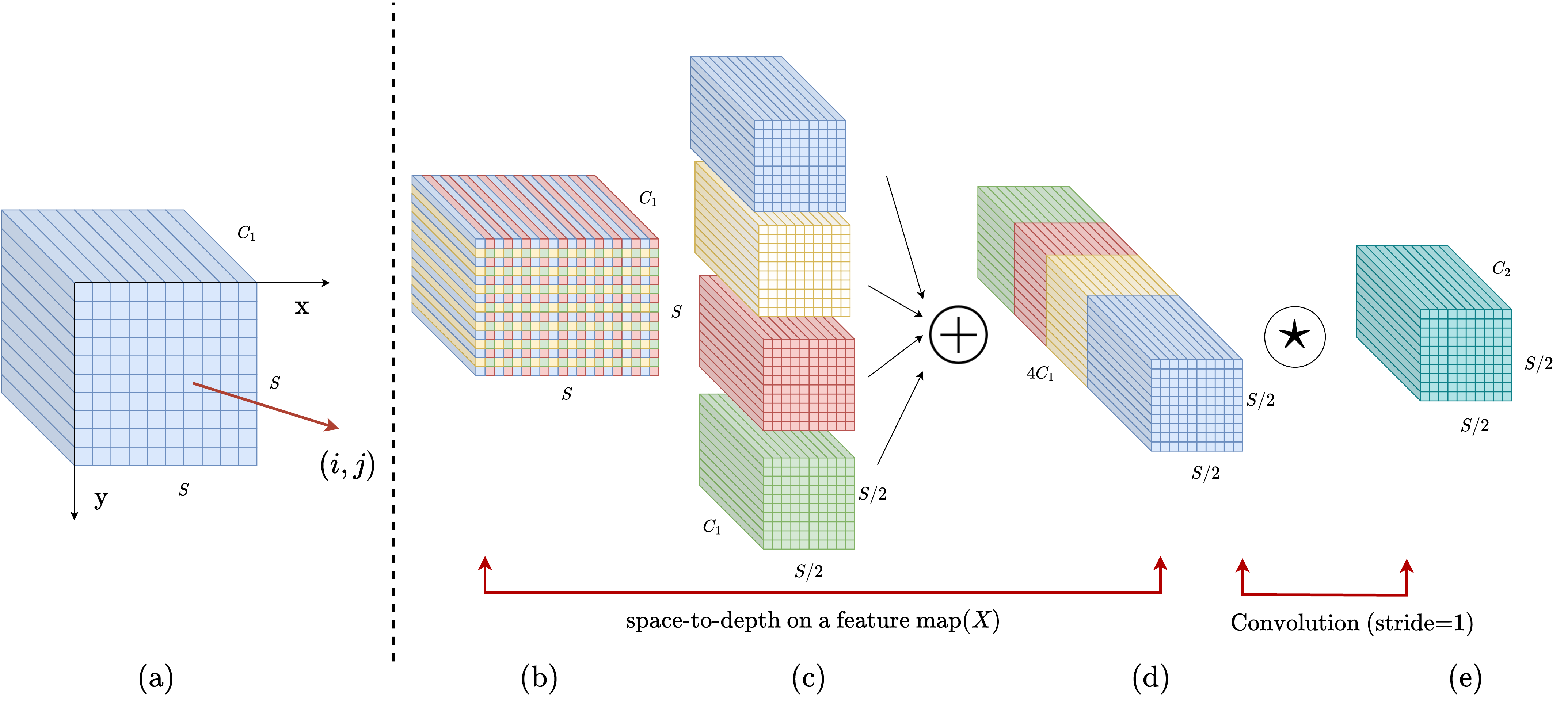}
    \caption{Illustration of SPD-Conv when $scale=2$ (see text for details).}
    \label{fig:spd}
\end{figure}

\section{A New Building Block: SPD-Conv}\label{sec:space-to-depth}

SPD-Conv is comprised of a Space-to-depth (SPD) layer followed by a non-strided convolution layer. This section describes it in detail. 

\subsection{Space-to-depth (SPD)}

Our SPD component generalizes a (raw) image transformation technique \cite{sajjadi2018frame} to downsampling feature maps inside and throughout a CNN, as follows.

Consider any intermediate feature map $X$ of size $S \times S \times C_1$, slice out a sequence of sub feature maps as

\noindent $f_{0,0} = X[0\mathbin{:}S\mathbin{:}scale, 0\mathbin{:}S\mathbin{:}scale], f_{1,0} = X[1\mathbin{:}S\mathbin{:}scale, 0\mathbin{:}S\mathbin{:}scale], \ldots, $ \\  $~\hspace{5cm} f_{scale\mathbin{-}1,0} = X[scale\mathbin{-}1\mathbin{:}S\mathbin{:}scale, 0\mathbin{:}S\mathbin{:}scale];$ \\ 
$f_{0,1} = X[0\mathbin{:}S\mathbin{:}scale, 1\mathbin{:}S\mathbin{:}scale], \; f_{1,1}, \;\; \ldots,$ \\
$~\hspace{5cm} f_{scale\mathbin{-}1,1} = X[scale\mathbin{-}1\mathbin{:}S\mathbin{:}scale, 1\mathbin{:}S\mathbin{:}scale];$ \\
$~\hspace{5.5cm}\vdots$ \\ 
$f_{0,scale\mathbin{-}1} = X[0\mathbin{:}S\mathbin{:}scale, scale\mathbin{-}1\mathbin{:}S\mathbin{:}scale], \; f_{1,scale\mathbin{-}1}, \;\; \ldots,$ \\
$~\hspace{3cm} f_{scale\mathbin{-}1,scale\mathbin{-}1} = X[scale\mathbin{-}1\mathbin{:}S\mathbin{:}scale, scale\mathbin{-}1\mathbin{:}S\mathbin{:}scale]$.\\
In general, given any (original) feature map $X$, a sub-map $f_{x,y}$ is formed by all the entries $X(i,j)$ that $i+x$ and $j+y$ are divisible by $scale$. Therefore, each sub-map downsamples $X$ by a factor of $scale$. Fig.~\ref{fig:spd}(a)(b)(c) give an example when $scale=2$, where we obtain four sub-maps $f_{0,0}, f_{1,0}, f_{0,1}, f_{1,1}$ each of which is of shape $(\frac{S}{2}, \frac{S}{2}, C_1)$ and downsamples $X$ by a factor of 2.

Next, we concatenate these sub feature maps along the channel dimension and thereby obtain a feature map $X'$, which has a reduced spatial dimension by a factor of $scale$ and an increased channel dimension by a factor of $scale^2$. In other words, SPD transforms feature map $X(S, S, C_1)$ into an intermediate feature map $X'(\frac{S}{scale}, \frac{S}{scale}, scale^2 C_1)$. Fig.~\ref{fig:spd}(d) gives an illustration using $scale=2$.

\subsection{Non-strided Convolution}

After the SPD feature transformation layer, we add a non-strided (i.e., stride=1) convolution layer with $C_2$ filters where $C_2 < scale^2 C_1$, and further transforms $X'(\frac{S}{scale}, \frac{S}{scale}, scale^2 C_1) \rightarrow X''(\frac{S}{scale}, \frac{S}{scale}, C_2)$. The reason we use non-strided convolution is to retain all the discriminative feature information as much as possible. Otherwise, for instance, using a 3 $\times$ 3 filer with stride=3, feature maps will get ``shrunk'' yet each pixel is sampled only once; if stride=2, {\em asymmetric sampling} will occur where even and odd rows/columns will be sampled different times. In general, striding with a step size greater than 1 will cause {\em non-discriminative loss} of information although at the surface, it appears to convert feature map $X(S, S, C_1)\rightarrow X''(\frac{S}{scale},\frac{S}{scale},C_2)$ too (but without $X'$).

\section{How to Use SPD-Conv: Case Studies}

To explain how to apply our proposed method to redesigning CNN architectures, we use two most representative categories of computer vision models: object detection and image classification. This is without loss of generality as almost all CNN architectures use strided convolution and/or pooling operations to downsample feature maps.

\subsection{Object Detection}

YOLO is a series of very popular object detection models, among which we choose the latest YOLOv5~\cite{yolov5} to demonstrate. YOLOv5 uses CSPDarknet53~\cite{bochkovskiy2020yolov4} with a SPP~\cite{he2015spatial} module as its backbone, PANet~\cite{Liu_2018_CVPR} as its neck, and the YOLOv3 head~\cite{redmon2018yolov3} as its detection head. In addition, it also uses various data augmentation methods and some modules from YOLOv4~\cite{bochkovskiy2020yolov4} for performance optimization. It employs the cross-entropy loss with a sigmoid layer to compute objectness and classification loss, and the CIoU loss function~\cite{zheng2020distance} for localization loss. The CIoU loss takes more details than IoU loss into account, such as edge overlapping, center distance, and width-to-height ratio.

\begin{figure}[t]
    \centering
    \includegraphics[width=0.85 \linewidth]{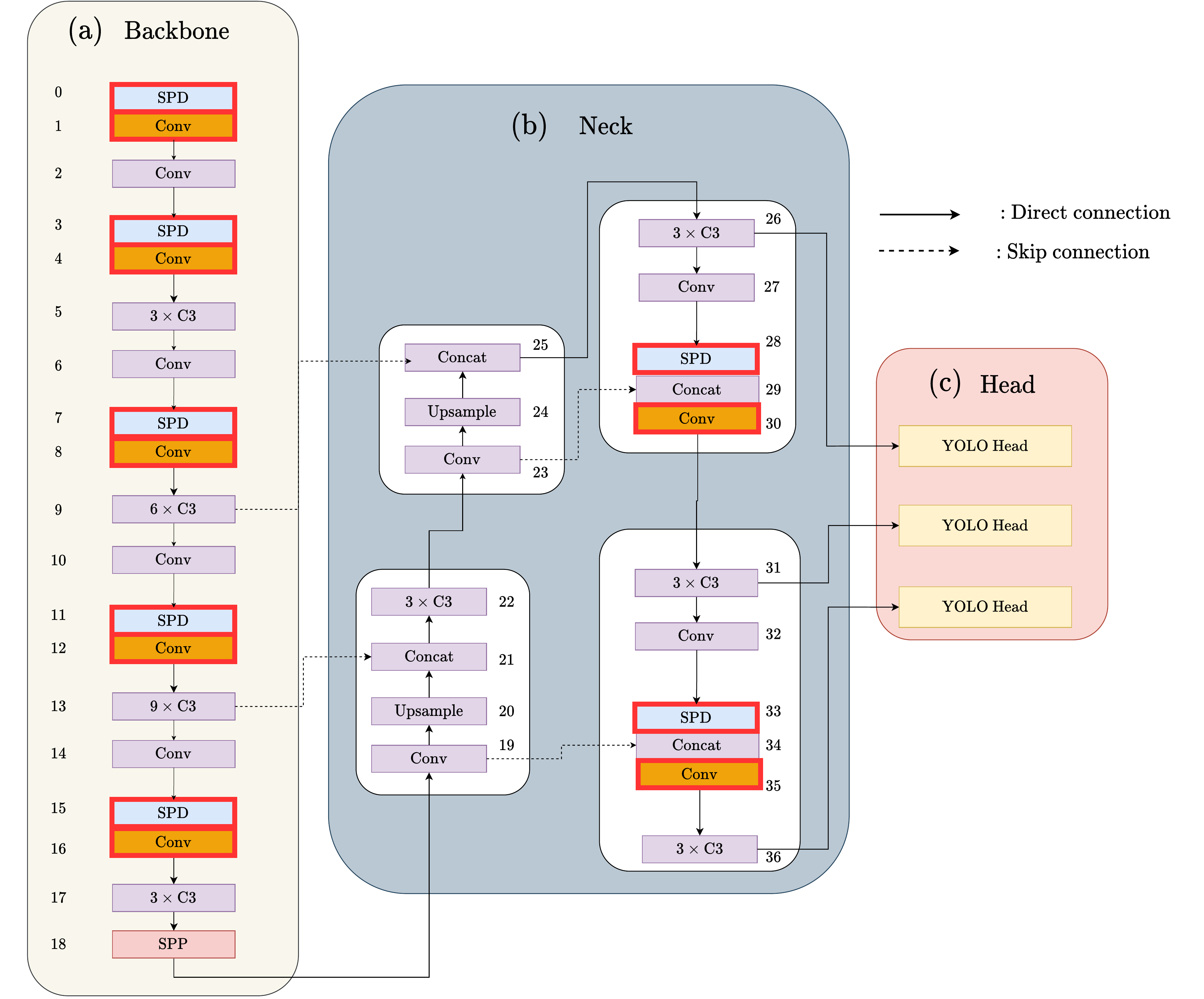}
    \caption{Overview of our YOLOv5-SPD. Red boxes are where the replacement happens.}
    \label{fig:yolov5_sd}
\end{figure}

{\bf YOLOv5-SPD.} We apply our method described in Section~\ref{sec:space-to-depth} to YOLOv5 and obtain YOLOv5-SPD (Fig.~\ref{fig:yolov5_sd}), simply by replacing the YOLOv5 stride-2 convolutions with our SPD-Conv building block. There are 7 instances of such replacement because YOLOv5 uses five stride-2 convolution layers in the backbone to downsample the feature map by a factor of $2^5$, and two stride-2 convolution layers in the neck. There is a concatenation layer after each strided convolution in YOLOv5 neck; this does not alter our approach and we simply keep it between our SPD and Conv.

{\bf Scalability.} YOLOv5-SPD can suit different application or hardware needs by easily scaling up and down in the same manner as YOLOv5. Specifically, we can simply adjust (1) the number of filters in every non-strided convolution layer and/or (2) the repeated times of C3 module (as in Fig. \ref{fig:yolov5_sd}), to obtain different versions of YOLOv5-SPD. The first is referred to as {\em width scaling} which changes the original width $n_w$ (number of channels) to $\lceil n_w \times width\_factor \rceil_8$ (rounded off to the nearest multiple of 8). The second is referred to as {\em depth scaling} which changes the original depth $n_d$ (times of repeating the C3 module; e.g., 9 as in $9\times \text{C3}$ in Fig. \ref{fig:yolov5_sd}) to $\lceil n_d \times depth\_factor \rceil$. 
This way, by choosing different width/depth factors, we obtain {\it nano, small, medium}, and {\it large} versions of YOLOv5-SPD as shown in Table \ref{tab:scaling}, 
where factor values are chosen the same as YOLOv5 for the purpose of comparison in our experiments later.

\begin{table}[h]
    \centering
    \caption{Scaling YOLOv5-SPD to obtain different versions that fit different use cases.}
    \begin{tabular}{lcc}
     \hline
 
     {Models} & {Depth\_Factor} & {Width\_Factor}  \\ \midrule
     
     YOLOv5-SPD-\textit{n} & 0.33  & 0.25 \\
     YOLOv5-SPD-\textit{s} & 0.33  & 0.50 \\
     YOLOv5-SPD-\textit{m} & 0.67  & 0.75 \\
     YOLOv5-SPD-\textit{l} & 1.00  & 1.00 \\
     \bottomrule
    \end{tabular}
    
    \label{tab:scaling}
\end{table}

\subsection{Image Classification}

A classification CNN typically begins with a stem unit that consists of a stride-2 convolution and a pooling layer to reduce the image resolution by a factor of four. A popular model is ResNet~\cite{he2016deep} which won the ILSVRC 2015 challenge. 
ResNet introduces residual connections to allow for training a network as deep as up to 152 layers. It also significantly reduces the total number of parameters by only using a single fully-connected layer. A softmax layer is employed at the end to normalize class predictions.

{\bf ResNet18-SPD and ResNet50-SPD.}
ResNet-18 and ResNet-50 both use a total 
number of four stride-2 convolutions and one max-pooling layer of stride 2 to downsample each input image by a factor of $2^5$. Applying our proposed building block, we replace the four strided convolutions with SPD-Conv; but on on the other hand, we simply remove the max pooling layer because, since our main target is low-resolution images, the datasets used in our experiments have rather small images ($64 \times 64$ in Tiny ImageNet and $32 \times 32$ in CIFAR-10) and hence pooling is unnecessary. For larger images, such max-pooling layers can still be replaced the same way by SPD-Conv. The two new architectures are shown in Table \ref{table:resnet}.

\input{tables/resnet_architecture}

\section{Experiments}

This section evaluates our proposed approach SPD-Conv using two representative computer vision tasks, object detection and image classification. 

\subsection{Object Detection}

\textbf{Dataset \& Setup.} We use the COCO-2017 dataset~\cite{coco} which is divided into {\tt train2017} (118,287 images) for training, {\tt val2017} (5,000 images; also called {\tt minival}) for validation, and {\tt test2017} (40,670 images) for testing. We use a wide range of state-of-the-art baseline models as listed in Tables \ref{tab:validation_table} and \ref{tab:testdev}. We report the standard metric of average precision (AP) on {\tt val2017} under different IoU thresholds [0.5:0.95] and object sizes (small, medium, large). We also report the AP metrics on {\tt test-dev2017} (20,288 images) which is a subset of {\tt test2017} with accessible labels. However, the labels are not publicly released but one needs to submit all the {\em predicted} labels in JSON files to the {\tt CodaLab COCO Detection Challenge} \cite{codalab} to retrieve the evaluated metrics, which we did.

\textbf{Training.} We train different versions (nano, small, medium, and large) of YOLOv5-SPD and all the baseline models on {\tt train2017}. Unlike most other studies, we {\em train from scratch without using transfer learning}. This is because we want to examine the {\em true learning capability} of each model without being disguised by the rich feature representation it inherits via transfer learning from ideal (high quality) datasets such as ImageNet.   This was carried out on our own models ($*$-SPD-n/s/m/l) and all the existing YOLO-series models (v5, X, v4, and their scaled versions like nano, small, large, etc.). The other baseline models still used transfer learning because of our lack of resource (training from scratch consumes an enormous amount of GPU time). However, note that this simply means that {\em those baselines are placed in a much more advantageous position} than our own models as they benefit from high quality datasets.

We choose the SGD optimizer with momentum 0.937 and a weight decay of 0.0005. The learning rate linearly increases from 0.0033 to 0.01 during three warm-up epochs, followed by a decrease using the Cosine decay strategy to a final value of  0.001. The {\it nano} and {\it small} models are trained on four V-100 32 GB GPU with a batch size of 128, while {\it medium} and {\it large} models are trained with batch size 32. CIoU loss~\cite{zheng2020distance} and cross-entropy loss are adopted for objectness and classification. We also employ several data augmentation techniques to mitigate overfitting and improve performance for {\em all} the models; these techniques include (i) photometric distortions of hue, saturation, and value, (ii) geometric distortions such as translation, scaling, shearing, fliplr and flipup, and (iii) multi-image enhancement techniques such as mosaic and cutmix. Note that augmentation is not used at inference. The hyperparameters are adopted from YOLOv5 without re-tuning.

\input{tables/table1}
\input{tables/table2}

\subsubsection{Results}
~\par

Table \ref{tab:validation_table} reports the results on {\tt val2017} and Table \ref{tab:testdev} reports the results on {\tt test-dev}. The $\text{AP}_{\text{S}}, \text{AP}_{\text{M}}, \text{AP}_{\text{L}} $ in both tables mean the AP for small/medium/ large {\em objects}, which should not be confused with {\em model} scales (nano, small, medium, large). The image resolution $640 \times 640$ as shown in both tables is not considered high in object detection (as opposed to image classification) because the resolution on the actual objects is much lower, especially when the objects are small.

\textbf{Results on val2017.}
Table \ref{tab:validation_table} is organized by model scales, as separated by horizontal lines (the last group are large-scale models). In the first category of nano models, our YOLOv5-SPD-\textit{n} is the best performer in terms of both AP and AP$_{\text{S}}$: its AP$_{\text{S}}$ is 13.15\% higher than the runner-up, YOLOv5n, 
and its overall AP is 10.7\% higher than the runner-up, also YOLOv5n.

In the second category, small models, our YOLOv5-SPD-\textit{s} is again the best performer on both AP and AP$_{\text{S}}$, although this time YOLOX-S is the second best on AP.

In the third, medium model category, the AP performance gets quite close although our YOLOv5-SPD-\textit{m} still outperforms others. On the other hand, our AP$_{\text{S}}$ has a larger winning margin (8.6\% higher) than the runner-up, which is a good sign because SPD-Conv is especially advantageous for smaller objects and lower resolutions.

Lastly for large models, YOLOX-L achieves the best AP while our YOLOv5-SPD-\textit{l} is only slightly (3\%) lower (yet much better than other baselines shown in the bottom group). On the other hand, our AP$_{\text{S}}$ remains the highest, which echos SPD-Conv's advantage mentioned above.

\textbf{Results on test-dev2017.}
As presented in Table \ref{tab:testdev}, our YOLOv5-SPD-\textit{n} is again the clear winner in the nano model category on AP$_{\text{S}}$, with a good winning margin (19\%) over the runner-up, YOLOv5n. For the average AP, although it appears as if EfficientDet-D0 performed better than ours, that is because EfficientDet has almost double parameters than ours and was trained using high-resolution images (via transfer learning, as indicated by ``Trf'' in the cell) and AP is highly correlated with resolution. This training benefit is similarly reflected in the small model category too.

In spite of this benefit that other baselines receive, our approach reclaims its top rank in the next category, medium models, on both AP and AP$_{\text{S}}$. Finally in the large model category, our YOLOv5-SPD-\textit{l} is also the best performer on AP$_{\text{S}}$, and closely matches YOLOX-L on AP.

\textbf{Summary.} 
It is clear that, by simply replacing the strided convolution and pooling layers with our proposed SPD-Conv building block, a neural net can significantly improves its accuracy, while maintaining the same level of parameter size. The improvement is more prominent when objects are small, which meets our goal well. Although we do not constantly notch the first position in all the cases, SPD-Conv is the only approach that {\em consistently} performs very well; it is only occasionally a (very close) runner-up if not performing the best, and is {\em always} the winner on AP$_{\text{S}}$ which is the chief metric we target.

Lastly, recall that we have adopted YOLOv5 hyperparameters without re-tuning, which means that our models will likely perform even better after dedicated hyperparameter tuning. Also recall that all the non-YOLO baselines (and PP-YOLO) 
were trained using transfer learning and thus have benefited from high quality images, while ours do not.

\textbf{Visual comparison.}
For a visual and intuitive understanding, we provide two real examples using two randomly chosen images, as shown in Fig.~\ref{fig:object_comparison}. We compare YOLOv5-SPD-\textit{m} and YOLOv5m since the latter is the best performer among all the baselines in the corresponding (medium) category. 
Fig.~\ref{fig:object_comparison}(a)(b) demonstrates that YOLOv5-SPD-\textit{m} is able to detect the occluded giraffe which YOLOv5m misses, and 
Fig.~\ref{fig:object_comparison}(c)(d) shows that YOLOv5-SPD-\textit{m} detects very small objects (a face and two benches) while YOLOv5m fails to.

\input{tables/plot9}

\subsection{Image Classification}

\textbf{Dataset \& Setup.} For the task of image classification, we use the Tiny ImageNet \cite{le2015tiny} and CIFAR-10 datasets~\cite{cifar10}. Tiny ImageNet is a subset of the ILSVRC-2012 classification dataset and contains 200 classes. Each class has 500 training images, 50 validation images, and 50 test images. Each image is of resolution $64 \times 64 \times 3$ pixels.  
CIFAR-10 consists of 60,000 images of resolution $32 \times 32 \times 3$, including 50,000 training images and 10,000 test images. There are 10 classes with 6,000 images per class. We use the top-1 accuracy as the metric to evaluate the classification performance.

\begin{figure}[h]\vspace{-7mm}
    \centering
    \includegraphics[width=.9\linewidth]{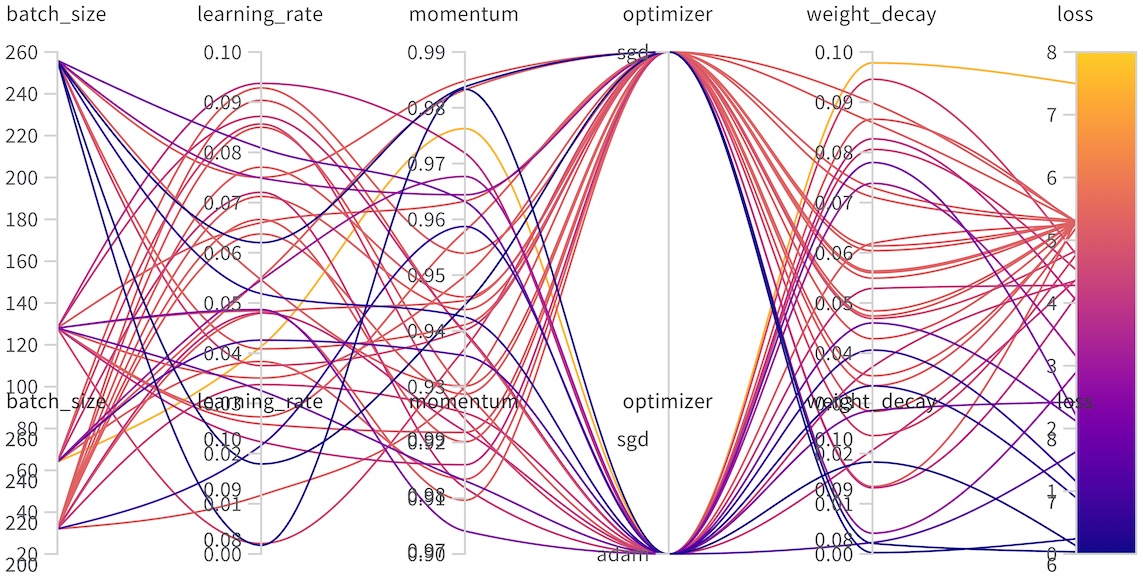}
    \caption{\footnotesize Hyperparameter tuning in image classification: a sweep plot using {\tt wandb}.}
    \label{fig:hyperparameter}
\end{figure}

\textbf{Training.} We train our ReseNet18-SPD model on Tiny ImageNet. We perform random grid search to tune hyperparameters including learning rate, batch size, momentum, optimizer, and weight decay. Fig.~\ref{fig:hyperparameter} shows a sample hyperparameter sweep plot generated using the {\tt wandb} MLOPs. The outcome is the SGD optimizer with a learning rate of 0.01793 and momentum of 0.9447, a mini batch size of 256, weight decay regularization of 0.002113, and 200 training epochs. Next, we train our ResNet50-SPD model on CIFAR-10. The hyperparameters are adopted from the ResNet50 paper, where SGD optimizer is used with an initial learning rate 0.1 and momentum 0.9, batch size 128, weight decay regularization 0.0001, and 200 training epochs. For both ReseNet18-SPD and ReseNet50-SPD, we use the same decay function as in ResNet to decrease the learning rate as the number of epochs increases.
\vspace{-7mm}

\textbf{Testing.} The accuracy on Tiny ImageNet is evaluated on the validation dataset because the ground truth in the test dataset is not available. The accuracy on CIFAR-10 is calculated on the test dataset. 

{\it \bf Results.} Table \ref{tab:classification_table} summarizes the results of top-1 accuracy. It shows that our models, ResNet18-SPD and ResNet50-SPD, clearly outperform all the other baseline models.
\input{tables/tiny_imagenet1}

\begin{figure}[h]\vspace{-2mm}
    \centering
    \includegraphics[width=0.9 \linewidth]{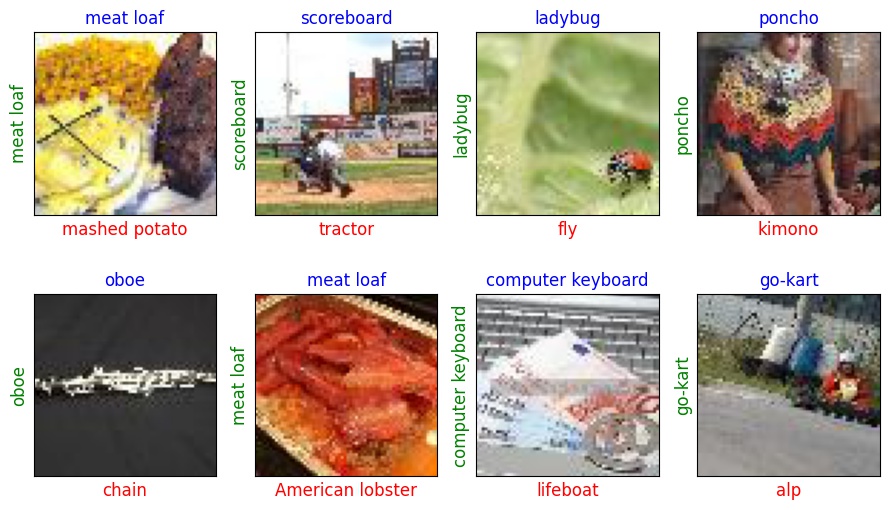}\vspace{-3mm}
    \caption{Green labels: ground truth. Blue labels: ResNet18-SPD predictions. Red labels: ResNet-18 predictions.}
    \label{fig:my_label2}\vspace{-5mm}
\end{figure}

Finally, we provide in Fig.~\ref{fig:my_label2} a visual illustration using Tiny ImageNet. It shows 8 examples misclassified by ResNet18 and correctly classified by ResNet18-SPD. The common characteristics of these images is that the resolution is low and therefore presents a challenge to the standard ResNet which loses fine-grained information during its strided convolution and pooling operations.

\section{Conclusion}

This paper identifies a common yet defective design in existing CNN architectures, which is the use of strided convolution and/or pooling layers. It will result in the loss of fine-grained feature information especially on low-resolution images and small objects. We then propose a new CNN building block called SPD-Conv that eliminates the strided and pooling operations altogether, by replacing them with a space-to-depth convolution followed by a non-strided convolution. This new design has a big advantage of downsampling feature maps while retaining the discriminative feature information. It also represents a general and unified approach that can be easily applied to perhaps any CNN architecture and to strided conv and pooling the same way. We provide two most representative use cases, object detection and image classification, and demonstrate via extensive evaluation that SPD-Conv brings significant performance improvement on detection and classification accuracy. We anticipate it to widely benefit the research community as it can be easily integrated into existing deep learning frameworks such as PyTorch and TensorFlow.

\bibliographystyle{splncs04}
\bibliography{ref}

\end{document}

%% file: tables/comparison_plot.tex
\begin{figure}[t]
    \centering
    \subfloat[Nano, small, and medium models.]{\includegraphics[width=0.5\linewidth]{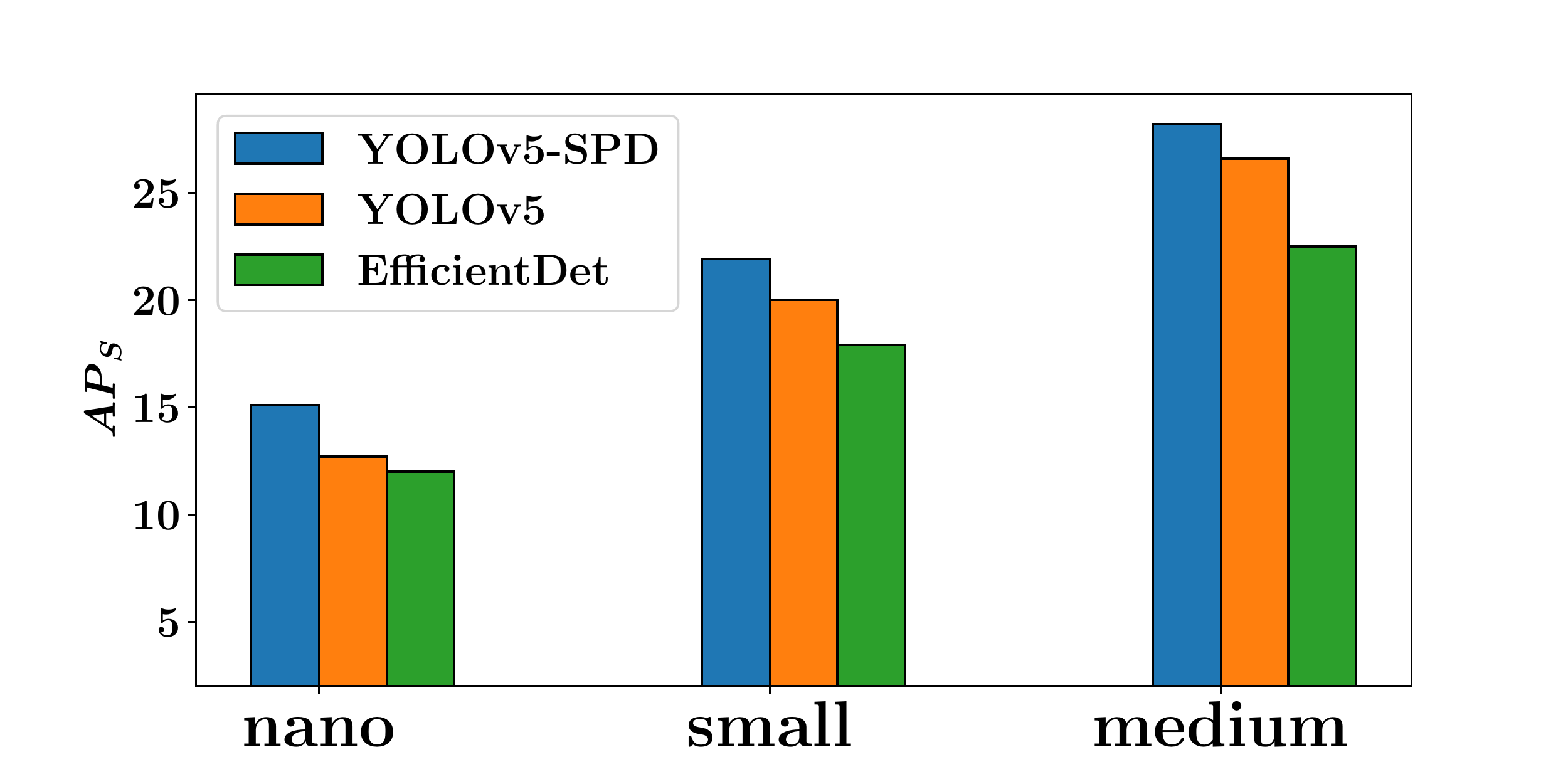} }
    \subfloat[Large-scale models.]{\includegraphics[width=0.5\linewidth]{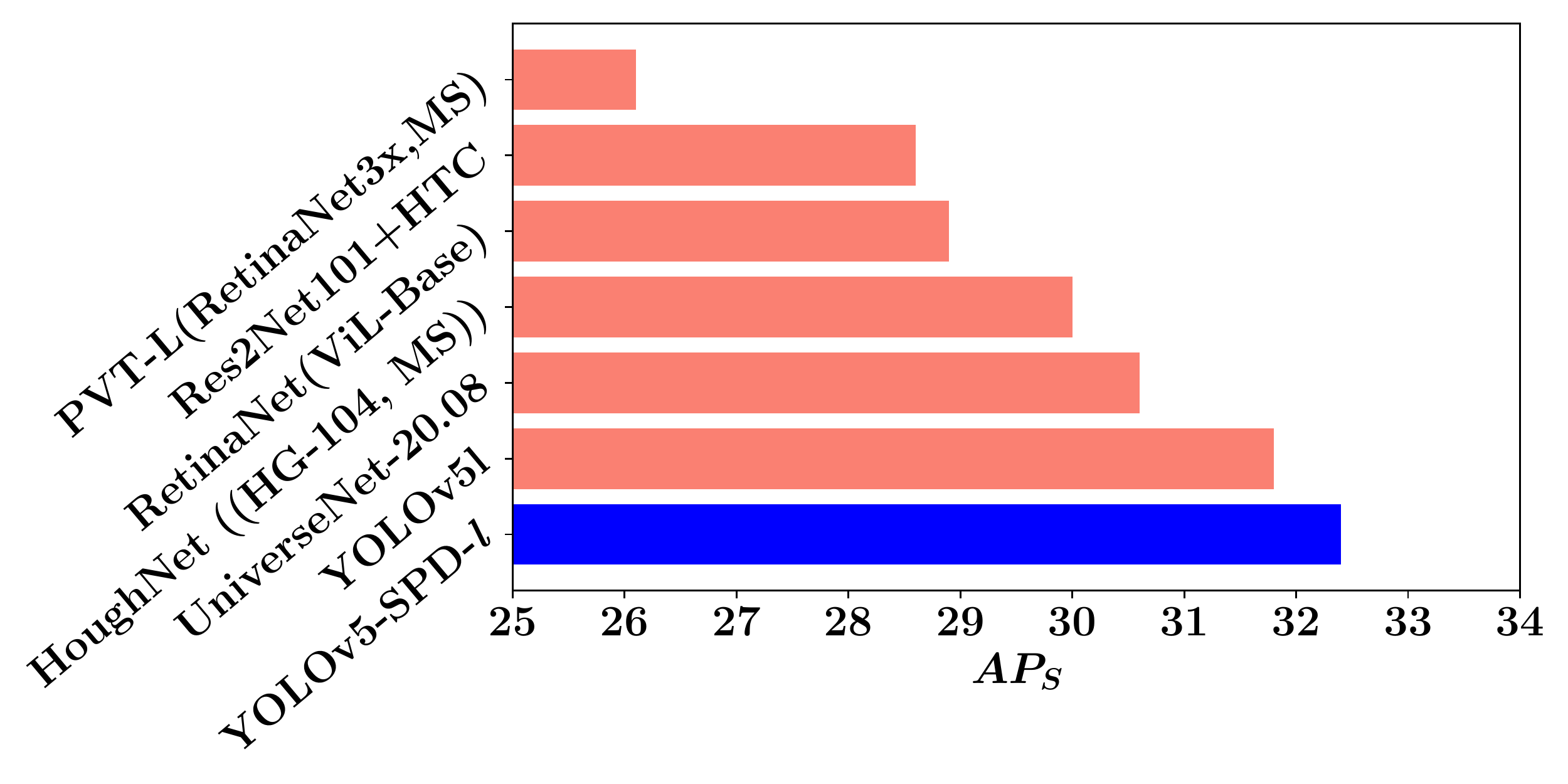} }
    \caption{Comparing AP for small objects ($AP_{S}$). ``SPD'' indicates our approach.}
    \label{fig:example}
\end{figure}

%% file: tables/try.tex
\begin{table}[t]
\begin{minipage}[t]{0.46\linewidth}
\caption{\footnotesize A taxonomy of OD models.}
\label{table:cls}
\fontsize{6.8}{10}\selectfont
\centering
      \begin{tabular}{|c|c|c|}
        \toprule

        {Model} & {Anchor-based} & {Anchor-free} \\ \midrule

        One-stage & \makecell{
        Faster R-CNN~\cite{ren2015faster},\\
        SSD~\cite{liu2016ssd},\\
        RetinaNet~\cite{lin2017focal},\\
        EfficientDet~\cite{tan2020efficientdet}, \\
        YOLO~\cite{redmon2018yolov3,bochkovskiy2020yolov4,yolov5,wang2021scaled}
        } & 
             \makecell{FCOS~\cite{tian2019fcos},\\
             CenterNet~\cite{duan2019centernet},\\
            DETR~\cite{carion2020end}, \\
            YOLOX~\cite{ge2021yolox}} \\ \midrule
        
             Two-stage & \makecell{R-CNN~\cite{girshick2014rich},\\
             Fast R-CNN~\cite{girshick2015fast}}
             & \makecell{  RepPoints, \\
             CenterNet2}\\
       
             \bottomrule
        \end{tabular}

\end{minipage} \hfill
\begin{minipage}[t]{0.54\linewidth}

\centering
      \captionof{figure}{\footnotesize A one-stage object detection pipeline.}
      \includegraphics[trim=8mm 0 0 2mm,clip,width=\linewidth]{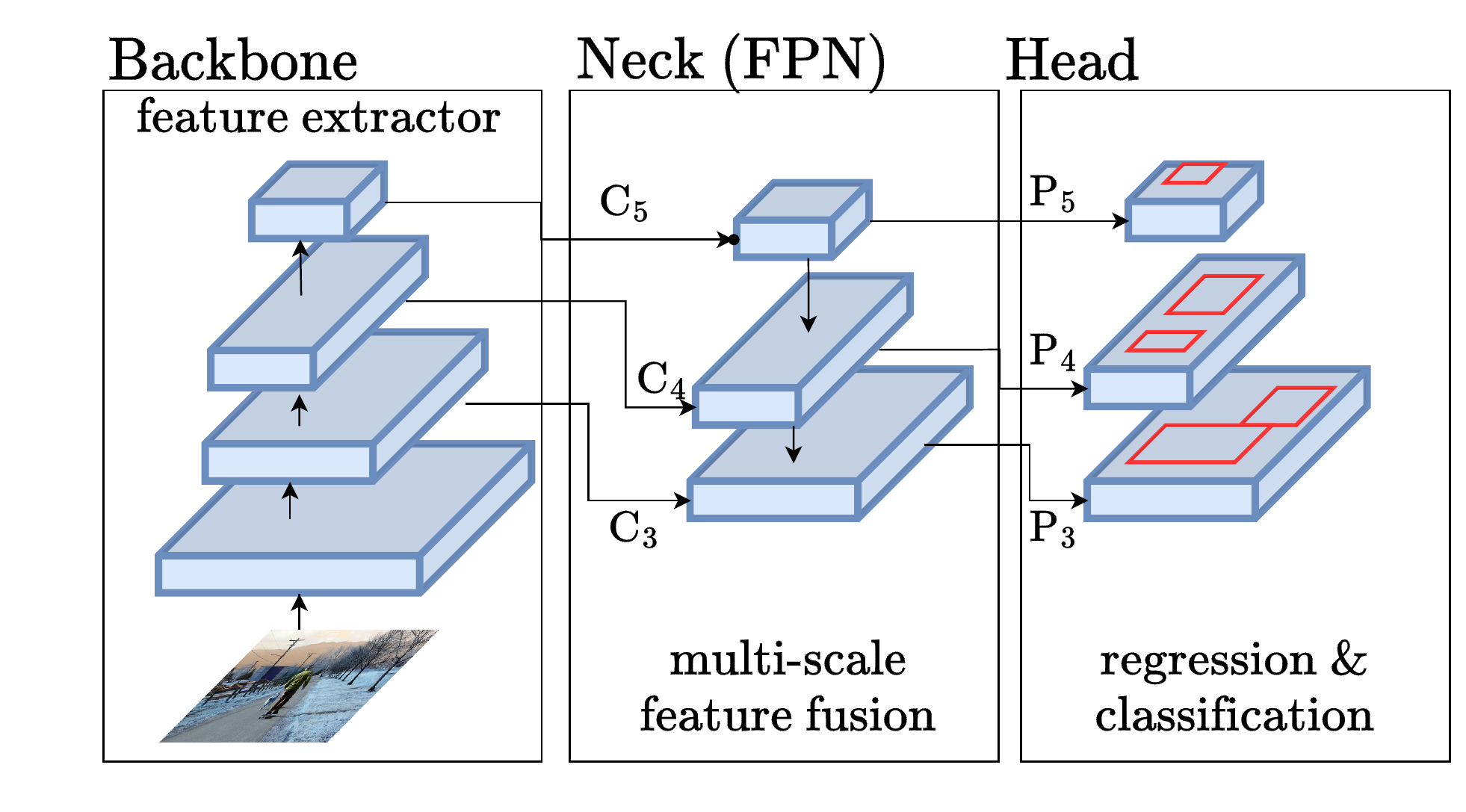}

    \label{fig:object_de}
\end{minipage}

\end{table}

%% file: tables/resnet_architecture.tex
\begin{table}[t]
\centering

\caption{Our ResNet18-SPD and ResNet50-SPD architecture.}

\begin{tabular}{c|c|c}
 \hline
 
 \textbf{\; Layer Name\;} & \textbf{$\quad$ ResNet18-SPD $\quad$} & \textbf{\; ResNet50-SPD \;}\\
 \hline
 spd1   & \multicolumn{2}{c}{\bf SPD-Conv}\\
 
 \hline
 conv1 & \multicolumn{2}{c}{$3\times 3$ kernel, 64 output channels} \\
 
 \hline

 conv2& $\begin{bmatrix} 3\times3, 64 \\ 3\times3, 64 \end{bmatrix} \times 2 $ & 
  $\begin{bmatrix} 1\times1, 64 \\ 3\times3, 64 \\ 1\times1, 256 \end{bmatrix} \times 3 $ \\ 
 
 \hline
 
 spd2   & \multicolumn{2}{c}{\bf SPD-Conv}\\
 
 \hline
 conv3& $\begin{bmatrix} 3\times3, 128 \\ 3\times3, 128 \end{bmatrix}  \times 2 $& $\begin{bmatrix} 1\times1, 128 \\ 3\times3, 128 \\ 1\times1, 512 \end{bmatrix} \times 4 $\\
 
 \hline
 
spd3   & \multicolumn{2}{c}{\bf SPD-Conv}\\
  
  \hline

 conv4& $\begin{bmatrix} 3\times3, 256 \\ 3\times3, 256 \end{bmatrix} \times 2 $ &$ \begin{bmatrix} 1\times1, 256 \\ 3\times3, 256 \\ 1\times1, 1024 \end{bmatrix} \times 6 $\\
 
 \hline
 
spd4  & \multicolumn{2}{c}{\bf SPD-Conv}\\
  
  \hline 
 
conv5& $\begin{bmatrix} 3\times3, 512 \\ 3\times3, 512 \end{bmatrix} \times 2 $& $\begin{bmatrix} 1\times1, 512 \\ 3\times3, 512 \\ 1\times1, 2048 \end{bmatrix} \times 3$\\

 \hline
 
 fc (fully conn.) & \multicolumn{2}{c}{Global avg. pooling + fc(no. of classes) + softmax} \\

  \hline
\end{tabular}

\label{table:resnet}

\end{table}

%% file: tables/table1.tex
\begin{table}[t]
\caption{Comparison on MS-COCO validation dataset (val2017).}
\label{tab:validation_table}
\centering
    \resizebox{\linewidth}{!}{

    \begin{tabular}{lcc|c|c|ccc}

     \toprule

     \textbf{Model} & \textbf{Backbone} & \textbf{Image} &\textbf{AP} & \textbf{$\text{AP}_\text{S}$} & \textbf{Params} & {\textbf{Latency (ms)} } \\
     & & \textbf{size} & & {\bf (small obj.)} & {\bf (M)} & {(batch\_size=1)}
     \\ \midrule
    
    \textbf{YOLOv5-SPD-\textit{n}}  &- & $640 \times 640 $  & \textbf{31.0} & \textbf{16.0} (\textcolor{blue}{\textbf{+13.15\%}}) & 2.2   & 7.3\\
     YOLOv5n    &- & $640 \times 640 $  & 28.0 & 14.14 & 1.9    & 6.3  \\
     YOLOX-Nano &- & $640 \times 640  $ & 25.3 & - & 0.9    & -  \\ \midrule
    \textbf{YOLOv5-SPD-\textit{s}} &- &$640 \times 640 $  & \textbf{40.0} &  \textbf{23.5} (\textcolor{blue}{\textbf{+11.4\%}}) & 8.7    & 7.3   \\
     YOLOv5s    &-  &$640 \times 640 $  & 37.4 & 21.09 & 7.2   & 6.4  \\
     YOLOX-S    &-  &$640 \times 640$  & 39.6 & - & 9.0    & 9.8 \\ \midrule
    \textbf{YOLOv5-SPD-\textit{m}} &- &$640 \times 640 $  & \textbf{46.5} & \textbf{30.3} (\textcolor{blue}{\textbf{+8.6\%}}) & 24.6  & 8.4 \\
     YOLOv5m  &-  &$640 \times 640  $    & 45.4 & 27.9 & 21.2   & 8.2 \\
     YOLOX-M   &- &$640 \times 640   $  & 46.4 & - & 25.3   & 12.3  \\ \midrule
    \textbf{YOLOv5-SPD-\textit{l}}  &- &$640 \times 640 $  & {48.5} & \textbf{32.4} (\textcolor{blue}{\textbf{+1.8\%}}) & 52.7 & 10.3 \\
     YOLOv5l   &-  &$640 \times 640  $   & 49.0 & 31.8 & 46.5  & 10.1   \\
     YOLOX-L    &-  &$640 \times 640 $     & {\bf 50.0} & - & 54.2  & 14.5  \\

     \midrule
     
     Faster R-CNN & R50-FPN & -& 40.2 & 24.2 & 42.0  & - \\
     Faster R-CNN+ & R50-FPN & -&42.0 & 26.6 & 42.0  & - \\
     DETR & R50 & -& 42.0 & 20.5 & 41.0  & - \\

     DETR-DC5 & ResNet-101 &  $800 \times 1333$ & 44.9 & 23.7 & 60.0  & - \\

     RetinaNet & ViL-Small-RPB & $800 \times 1333$ &44.2 & 28.8 & 35.7  & - \\

     \bottomrule
    \end{tabular}
    }
    
\end{table}

%% file: tables/table2.tex
\begin{table}[t]
\caption{Comparison on MS-COCO test dataset (test-dev2017).}
\label{tab:testdev}
\resizebox{\linewidth}{!}{
\centering

\begin{tabular}{l|c|c|c|c|c|c|c|c} 

\toprule

\textbf{Model} & \textbf{ImgSize}  & \textbf{Params} & \textbf{AP} & \textbf{AP\textsubscript{50}} &  \textbf{AP\textsubscript{75}}&  \textbf{AP\textsubscript{S}}  & \textbf{AP\textsubscript{M}}&  \textbf{AP\textsubscript{L}} \\ 

 &  & {\bf (M)} &  &  & & {\bf (small obj.)} &  &  \\

\midrule

\textbf{YOLOv5-SPD-\textit{n}} &$640 \times 640$ & 2.2  & \textbf{30.4} & 48.7 &32.4&\textbf{15.1}(\textcolor{blue}{\textbf{+19\%}})&33.9&37.4 \\

YOLOv5n &$640 \times 640$ & 1.9&28.1 & 45.7 & 29.8 & 12.7 & 31.3 & 35.4 \\

EfficientDet-D0 &$512 \times 512$ & 3.9& 33.8(Trf) & 52.2 & 35.8 & 12.0 & 38.3 & 51.2 \\

\midrule

\textbf{YOLOv5-SPD-\textit{s}} &$640 \times 640$ & 8.7  & \textbf{39.7} & 59.1 &43.1&\textbf{21.9}(\textcolor{blue}{\textbf{+9.5\%}})
&43.9&49.1 \\

YOLOv5s &$640 \times 640$ & 7.2&37.1 & 55.7 & 40.2 & 20.0 & 41.5 & 45.2 \\
EfficientDet-D1 &$640 \times 640$ & 6.6&39.6 & 58.6 & 42.3 & 17.9 & 44.3 & 56.0 \\
EfficientDet-D2 &$768 \times 768$ & 8.1& 43.0(Trf) & 62.3 & 46.2 & 22.5(Trf) & 47.0 & 58.4 \\

\midrule

\textbf{YOLOv5-SPD-\textit{m}} &$640 \times 640$ & 24.6  & \textbf{46.6} & 65.2 & 50.8& \textbf{28.2}(\textcolor{blue}{\textbf{+6\%}})
& 50.9 & 57.1 \\
YOLOv5m &$640 \times 640$ &21.2& 45.5 & 64.0 & 49.7 & 26.6 & 50.0 & 56.6 \\
YOLOX-M &$640 \times 640$ & 25.3  & 46.4 & 65.4 & 50.6 & 26.3&51.0&59.9 \\
EfficientDet-D3 &$896 \times 896$ & 12.0&45.8 & 65.0 & 49.3 & 26.6 & 49.4 & 59.8 \\
SSD512& $512 \times 512$&36.1&28.8&48.5&30.3&-&-&- \\
\midrule

\textbf{YOLOv5-SPD-\textit{l}} &$640 \times 640$ & 52.7  & {48.8} & 67.1 & 53.0 & \textbf{30.0} & 52.9 & 60.5  \\ 
YOLOv5l &$640 \times 640$ &46.5& 49.0 & 67.3 & 53.3 & 29.9 & 53.4 & 61.3 \\
YOLOX-L &$640 \times 640$ &54.2 & {\bf 50.0} & 68.5 & 54.5 & 29.8 &54.5 &64.4 \\
YOLOv4-CSP &$640 \times 640$&52.9& 47.5&66.2&51.7&28.2&51.2&59.8 \\
PP-YOLO &$608 \times 608$&52.9& 45.2& 65.2& 49.9& 26.3& 47.8&57.2 \\ 

 \midrule
 
YOLOX-X &$640 \times 640$ & 99.1  & 51.2 & 69.6 & 55.7 & 31.2  &56.1 &66.1\\ 
YOLOv4-P5 &$896 \times 896$&70.8& 51.8&70.3&56.6&33.4&55.7&63.4 \\
YOLOv4-P6 &$1280 \times 1280$&127.6& 54.5& 72.6& 59.8& 36.8& 58.3&65.9\\

RetinaNet      & $1280 \times 1280$ & 66.9 & 50.7 & 70.4 & 54.9 & 33.6 & 53.9 & 62.1 \\ 
(w/ SpineNet-143) &  &  &  &  & &  &  &  \\ 

\bottomrule

\end{tabular}

}

\end{table}

%% file: tables/plot9.tex
\begin{figure}[t]
    \centering
    \subfloat[Purple boxes: YOLOv5m predictions.]{\includegraphics[width=0.5\linewidth]{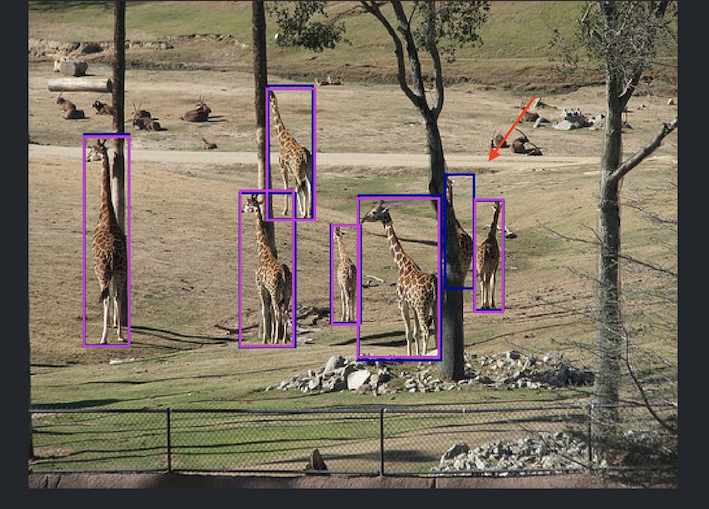} }
    \subfloat[Green boxes: YOLOv5-SPD-\textit{m} predictions.]{\includegraphics[width=0.5\linewidth]{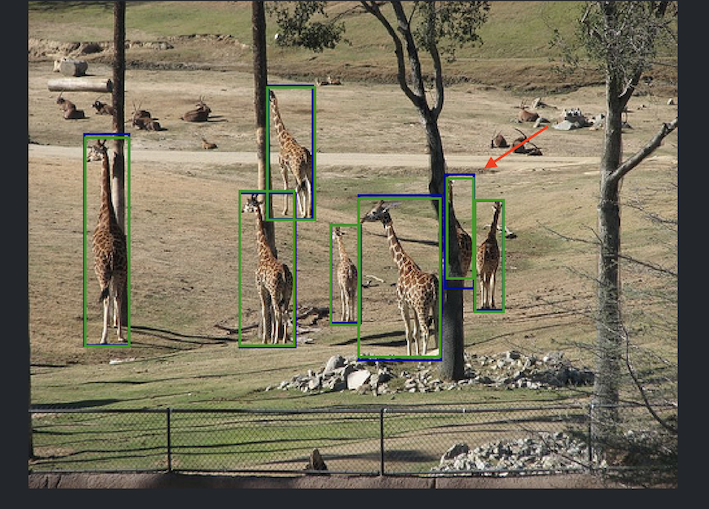} }
    \\
    \subfloat[Purple boxes: YOLOv5m predictions.]{\includegraphics[width=0.5\linewidth]{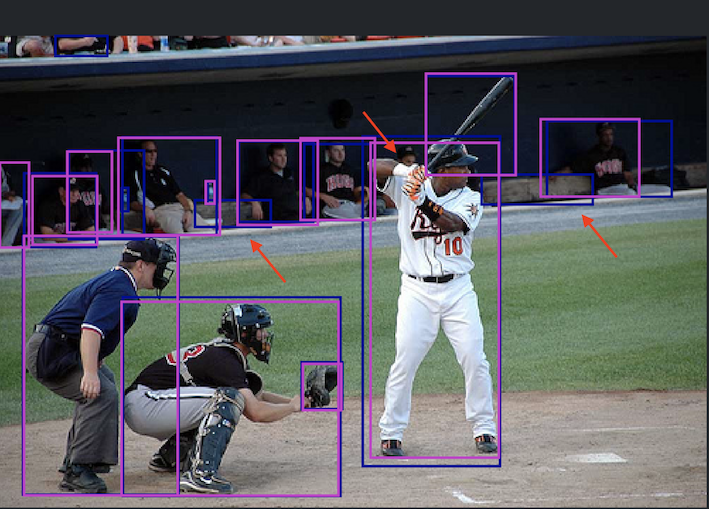} }
    \subfloat[Green boxes: YOLOv5-SPD-\textit{m} predictions.]{\includegraphics[width=0.5\linewidth]{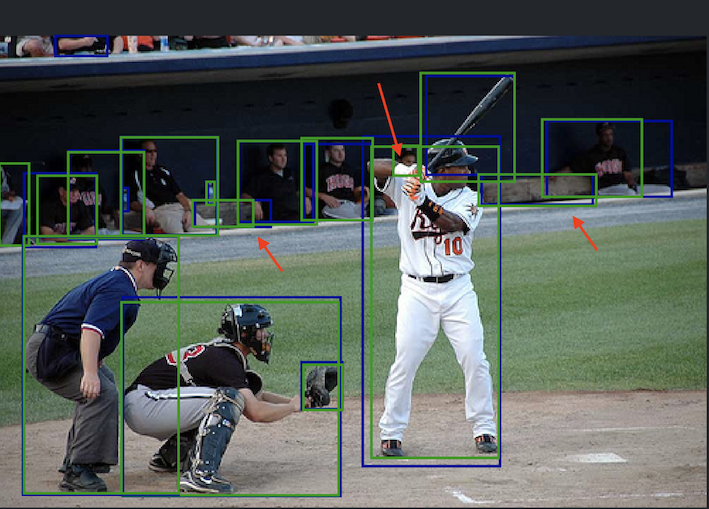} }

    \caption{Object detection examples from {\tt val2017}. Blue boxes indicate the ground truth. Red arrows highlight the differences.}
    
    \label{fig:object_comparison}
\end{figure}

%% file: tables/tiny_imagenet1.tex
\begin{table}[h]
    \caption{Image classification performance comparison.}
  \label{tab:classification_table}

    \centering

\resizebox{.87\linewidth}{!}{

    \begin{tabular}{l|c|c}

     \hline
     \hline

     {\textbf{Model}} &  \textbf{$\quad$ Dataset $\quad$}&{\textbf{Top-1 accuracy (\%)}}   \\ \midrule
     
    \textbf{ResNet18-SPD}    & Tiny ImageNet & \textbf{64.52}  \\
     
    ResNet18    & Tiny ImageNet& 61.68   \\

    Convolutional Nystromformer for Vision & Tiny ImageNet & 49.56 \\
    
    WaveMix-128/7 & Tiny ImageNet & 52.03  \\
    
     \midrule
    
    \textbf{ResNet50-SPD} & CIFAR-10 &\textbf{95.03}   \\
    ResNet50  & CIFAR-10 & 93.94 \\

    Stochastic Depth & CIFAR-10 & 94.77 \\
    Prodpoly & CIFAR-10 & 94.90 \\

     \bottomrule

    \end{tabular}
    
}
    
\end{table}